\documentclass[letterpaper,twocolumn,10pt,anonymous]{article}

\usepackage{usenix}
\usepackage{savesym}
\usepackage[T1]{fontenc}
\restoresymbol{TXF}{iint}
\usepackage{amsfonts}
\usepackage{amsmath}
\usepackage{textcomp}
\usepackage{bbding}
\usepackage{xspace}
\usepackage{pifont}
\usepackage[normalem]{ulem}
\usepackage{hyphenat}
\usepackage{algorithm}
\usepackage{algorithmicx}
\usepackage[noend]{algpseudocode}
\usepackage{mathtools}
\usepackage{xspace}
\usepackage{enumitem}
\usepackage{multirow}
\usepackage{svg}
\usepackage{balance}
\usepackage{titlesec}

\usepackage[most]{tcolorbox} 
\usepackage{import}

\titlespacing*{\section}{0pt}{4pt plus 2pt minus 1pt}{4pt plus 2pt minus 1pt}
\titlespacing*{\subsection}{0pt}{2pt plus 2pt minus 0pt}{2pt plus 2pt minus 0pt}

\setlist{nosep}

\newcommand{\sparagraph}[1]{\vspace{1mm}\noindent {\bf #1}\xspace}

\newif\ifremove
\removefalse

\begin{document}
\title{Hyperstroke: A Novel High-quality Stroke Representation for Assistive Artistic Drawing}

\author{
{\rm Haoyun Qin} \\ University of Pennsylvania
\and {\rm Jian Lin} \\ Saint Francis University
\and {\rm Hanyuan Liu} \\ City University of Hong Kong
\and {\rm Xueting Liu} \\ Saint Francis University
\and {\rm Chengze Li} \\ Saint Francis University
}

\maketitle

\begin{abstract}
Assistive drawing aims to facilitate the creative process by providing intelligent guidance to artists. Existing solutions often fail to effectively model intricate stroke details or adequately address the temporal aspects of drawing. We introduce hyperstroke, a novel stroke representation designed to capture precise fine stroke details, including RGB appearance and alpha-channel opacity. Using a Vector Quantization approach, hyperstroke learns compact tokenized representations of strokes from real-life drawing videos of artistic drawing. With hyperstroke, we propose to model assistive drawing via a transformer-based architecture, to enable intuitive and user-friendly drawing applications, which are experimented in our exploratory evaluation. 
\end{abstract}
\section{Introduction} \label{sec:intro}

Drawing is inherently an incremental process where artworks are created stroke-by-stroke, reflecting underlying drawing intent and locality. In this work, we investigate the problem of incremental drawing from the perspective of a drawing assistant. Our goal is to provide essential guidance to users in applying proper drawing strokes to complete visually pleasing artworks, considering the current unfinished canvas composition and the full or partial history of user strokes. Such an application enhances our understanding of the creative process and seamlessly integrates into existing artistic workflows in a suggest-then-accept manner.

\begin{figure}[!t]
  \centering
  \includegraphics[width=\linewidth]{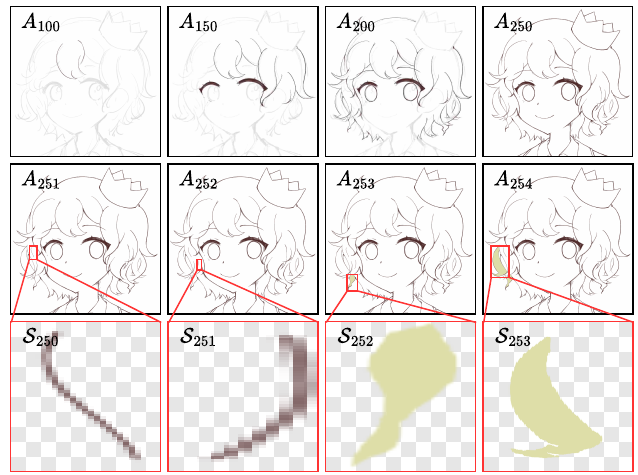}
  \caption{Example of real-life artistic drawing. The incremental drawing on canvas $A_t$ is recorded in the form of timelapse video. The user-provided stroke $\mathcal{S}_t$ is not included in the timelapse and has to be explicitly estimated.}
  \label{fig:timelapse}
\end{figure}

The existing literature focuses mainly on reproducing complete artworks using pre-defined stroke patterns~\cite{zheng2018strokenet, singh2021intelli, liu2021paint, huang2019learning} or performing incremental stroke prediction exclusively in the vector domain~\cite{ha2017neural, bhunia2020pixelor}. Recent diffusion-based models exhibit impressive results in the generation of artwork, but their generation must be performed in a single pass~\cite{rombach2022high, nitzan2024lazy}. This hinders iterative refinement and co-creation, which are essential in the drawing process.
We hypothesize that existing approaches may prioritize overall visuals but neglect the importance of strokes, which are the fundamental basic units contributing to an artwork in both spatial and temporal domains. This oversight is particularly detrimental for a drawing assistant. 
In Figure~\ref{fig:timelapse}, we illustrate several steps in which the user applies strokes. Real-life drawing of strokes is far more complex than simple shape primitives, involving specific movements, shape and color variations, etc. More importantly, these strokes exhibit opacity, i.e. \emph{alpha}, to blend additively over the canvas, crafting delicate details and shadings. Therefore, understanding and modeling strokes are crucial for modeling a drawing assistant.

In this work, we propose \emph{hyperstroke}, an efficient and expressive stroke representation to better model strokes in real-life artistic drawing with alpha-channel opacity. Our key insight is to employ a VQ-based model to learn a compact tokenized representation of grounded strokes within their bounding boxes. Our experiments demonstrate the efficiency of the hyperstroke design and, more importantly, show the potential to learn predictive incremental drawing under the hyperstroke formulation, using an encoder-decoder transformer architecture. We summarize our contributions as follows:

\begin{itemize}[]
    \item We introduce a novel stroke representation, \textit{hyperstroke}, to model delicate artistic drawing stroke appearance and opacity;
    \item We propose an updated VQGAN architecture to learn hyperstroke tokenization from real-life incremental drawing;
    \item We investigate to use transformer models to learn hyperstroke sequences for assistive incremental drawing.
   
\end{itemize}

\section{Method} 
\label{sec:method}

\begin{figure*}[!t]
  \centering
  \includegraphics[width=\linewidth]{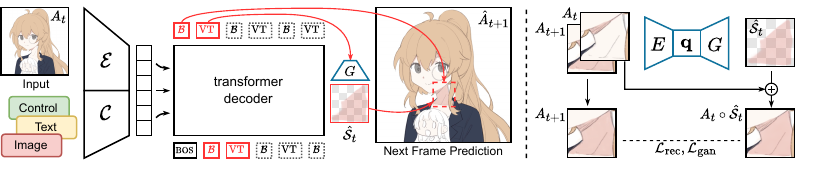}
  \caption{Overview of our framework. The right demonstrates the learning of tokenization in hyperstrokes (Section \ref{subsec:hyperstroke}), while the left shows our systematic design in predictive incremental drawing (Section \ref{subsec:sequence}). 
  }
  \label{fig:arch}
\end{figure*}

\subsection{Hyperstroke}
\label{subsec:hyperstroke}

\subsubsection{Formulation}

In this work, we introduce the novel \emph{hyperstroke} representation for modelling the strokes in practical artistic drawing. Unlike traditional methods that represent storkes as simple elliptical pixels, or vector primitives, our approach aims to capture the essence of real-life strokes with diverse appearances and alpha variations. By investigating the artistic drawing process, we observe several key properties within a stroke:

\noindent \textbf{Property 1: Independence in Representation.}
Strokes are additive in nature, meaning each new stroke is an additional layer alpha-blended onto the existing canvas, as we can observe in the strokes $\mathcal{S}$ of Figure~\ref{fig:timelapse}. Thus, the representation of a single stroke should be independent of the underlying canvas (i.e., all other strokes), regardless of their composition or color usage.

\noindent \textbf{Property 2: Spatial Sparsity.}
Strokes are inherently spatially sparse. During the drawing process, although the canvas may be extensive, each stroke is either detailed and confined to a small area or spans a larger region but is relatively coarse. Therefore, when extracted and normalized to a consistent scale, each stroke should carry a similar amount of low-scale information.

Based on these assumptions, we design our hyperstroke representation to be atomic and compact. Leveraging the sparsity property of strokes, we propose using bounding boxes to locate each stroke and encode only the pixels within them, for better expressiveness of strokes in smaller regions. Formally, we define the pixel-domain \textit{hyperstroke} \(\mathcal S = \langle \mathcal I, \mathcal B \rangle\), where \(\mathcal I = (I, \alpha)\) is a 4-channel alpha image and \(\mathcal B = (x_1, y_1, x_2, y_2)\) is the bounding box of $\mathcal I$. 
In this way, we can regard each stroke-box combo shown in the bottom two rows of Figure~\ref{fig:timelapse} as a hyperstroke.
When a hyperstroke \(\mathcal S\) is applied to an image \(A\), we denote the blending operation \(A \circ \mathcal S\) as:
\begin{equation}
    \label{eq:stroke-blend-def}
    \hspace{-1em}
    (A \circ \mathcal S) (x, y) = \begin{cases}
        \left ( I \cdot \alpha + A \cdot (1 - \alpha) \right ) (x, y) & 
        \begin{aligned}
            x_1 \leq x < x_2 \\
            y_1 \leq y < y_2
        \end{aligned}
        \\
        A(x,y). & \text{otherwise}
    \end{cases}
\end{equation}

\subsubsection{Tokenization}
To this point, we have formulated the hyperstrokes in the pixel space. However, this formulation proves ineffective for modeling incremental drawing, as learning pixel-domain hyperstrokes with temporal information is computationally intensive. Conversely, transformer models excel at modeling temporal sequences, which is more suitable for learning incremental drawing, suggesting the representation of hyperstrokes as discrete tokens. Specifically, we perform hyperstroke tokenization separately for $\mathcal I$ and $\mathcal B$. To tokenize bounding box $\mathcal{B}$, we first subdivide the image canvas into grids of \(C \times C\), with each grid cell having dimensions \(\lfloor W / C \rfloor \times \lfloor H / C \rfloor\), where \(W\) and \(H\) denotes width and height of the original canvas. For any bounding box $\mathcal B$, we compute its smallest exterior box that snaps to the grid and tokenize it in the form $\tilde{B} = (X_1, Y_1, X_2, Y_2) \in T_{\mathcal{B}}^4$, where the integer $X_1, X_1, Y_2, Y_2$ represents the indices of the grid corners to which the exterior box snaps, and $T_{\mathcal{B}}$ is a vocabulary of $\{0, 1, \ldots, C\}$. This grid-based design reduces the complexity of bounding box tokens without significantly compromising the encoding of the stroke image $\mathcal{I}$, using a slightly larger bounding box.

For the stroke pixels $\mathcal I$, we perform the same grid snapping strategy as \(\mathcal B\),
and then resize it to a consistent dimension \(W_T \times H_T\). We learn to tokenize its visual tokens $\tilde{\mathcal I} \in T_{\text{VT}}^k$ via a VQ-based approach, which will be explained in the following subsection.

\subsubsection{Training Hyperstroke from Real-life Incremental Drawing}
Tokenizing a 4-channel alpha image $\mathcal I$ appears straightforward due to existing standards such as VQGAN~\cite{esser2021taming}. However, we find the quality of the data contributing to visual token learning critical. Synthesizing arbitrary alpha strokes programmatically is one direction but would overcomplicate the final encoded tokens. Real-life strokes exhibit more specific distributions, as the drawing of each stroke follows human-specific aesthetic considerations. 
In this circumstance, sources recording practical human-drawn strokes with pixel-level opacity would be ideal for training, but such data is usually unavailable. Therefore, we attempt to collect strokes with alpha information from \emph{timelapse videos} (shown in Fig.~\ref{fig:timelapse}), which capture consecutive canvas frames whenever a new stroke is applied. Unfortunately, timelapse videos do not store any specific stroke information, so we have to estimate the strokes $\mathcal{S}_t$ from adjacent frame correspondences; but direct estimation is infeasible due to the ill-posed nature of inversing alpha blending. 
To address this, we propose an improved VQ model architecture to predict alpha strokes $\hat{S_t}$ from adjacent frames with implicit supervision, without requiring ground truth stroke.

We illustrate our VQ model design on the right of Figure~\ref{fig:arch}. The input is the concatenation $([A_t, A_{t+1}] \in \mathbb{R}^{H \times W \times 6})$ of any adjacent frames $A_t$ and $A_{t+1}$ in the data set. We use the encoder $E$ and a codebook $\mathcal{Z}$ to learn the tokenization of stroke features as $\mathbf{q}(E(A_t, A_{t+1}))$. We use a decoder $G$ to learn the reconstruction of the 4-channel stroke $\hat{S_t}$ from the learned tokens. Here, we supervise $\hat{S_t}$ by checking if $A_{t+1}$ can be obtained by blending $A_t$ with $\hat{S_t}$:
\begin{equation}
    \label{eq:vq-loss}
    \hspace{-1em}
    \begin{aligned}
        \mathcal L_{\text{VQ}} = 
        \mathcal L_{\text{rec}} \left ( A_{t+1}, \left ( A_t \circ \hat {\mathcal S_t} \right ) \right )
        &+ \left \| \operatorname{sg} \left [ E(A_t, A_{t+1}) \right ] - z_{\mathbf q} \right \|^2_2 \\
        &+ \left \| \operatorname{sg} \left [ z_{\mathbf q} \right ] - E(A_t, A_{t+1}) \right \|^2_2,
    \end{aligned}
\end{equation}
where $\mathcal L_{\text{rec}}$ is the sum of the MSE loss and the perceptual loss~\cite{zhang2018unreasonable} and the other two loss terms optimize the use of codebooks; $\operatorname{sg}[\cdot]$ denotes the stop-gradient operation. The blending operation $A_t \circ \hat {\mathcal S_t}$ in the supervision encourages the encoder $E$ to focus on a decoupled representation of the stroke, rather than memorizing $A_t$ and $A_{t+1}$. Besides, we also introduce adversarial learning with a discriminator $D$ for better decoder perceptual quality, with a similar implicit supervision:
\begin{equation}
    \label{eq:gan-loss}
    \mathcal L_{\text{GAN}} 
        = \left [
        \log D(A_{t+1}) + \log \left ( 1 - D \left ( A_t \circ \hat {\mathcal S_t} \right ) \right )
    \right ].
\end{equation}

\subsubsection{Data and Training Details}

We construct our dataset in two parts: a synthetic dataset and data from real-life timelapse videos.
For the synthetic data, we first perform a random crop of real artistic drawings. After that, we synthesize a Bezier stroke with varying widths and opacity and blend it with the cropped drawing to form the data. 
Since the synthetic data contains ground truth alpha for the stroke, we can use direct reconstruction loss \(\mathcal L_\text{rec}\) in Eq.~\ref{eq:vq-loss} instead of implicit supervision with additional alpha blending on the generator output. This direct supervision helps the model better understand opacity from the very beginning of training, thereby improving its learning capability on real-life data. 
Overall, our dataset consists of 85,425 synthetic data samples and 74,286 real data samples in the form of frame pairs. We mix the two types of supervision during training directly.

\subsection{Learning Drawing with Hyperstroke} \label{subsec:sequence}

Expanding on the stroke tokenization method outlined in Section \ref{subsec:hyperstroke}, we define incremental drawing as a sequence generation task, which can be effectively modeled with an encoder-decoder transformer model. The model, as shown on the left of Figure \ref{fig:arch}, leverages the encoder \(\mathcal E\), a Vision Transformer (ViT) \cite{dosovitskiy2020image}, to extract contextual information $\tau_c$ from the current canvas $A_t$. Furthermore, we use the CLIP model \cite{radford2021learning} \(\mathcal C\) to encode the guidance $\tau_g$ of controlling signals such as reference images and text descriptions.
We combine $\tau_c$ and $\tau_g$ embeddings and send them to the decoder through cross-attention, to predict subsequent \textit{hyperstroke} tokens $\left ( (\tilde{\mathcal B}, \tilde{\mathcal {I}}) \in T_{\mathcal B}^4 \times T_{\text{VT}}^k \right ) ^ n$ in an autoregressive manner, where \(k\) is the number of visual tokens for each stroke, and \(n\) is the number of \textit{hyperstrokes} to be predicted. With the VQ decoder model $G$, we will be able to decode and composite each hyperstroke back into the pixel domain, to form future frames from $A_{t+1}$ to $A_{t+n}$.

\begin{figure*}[t]
  \centering
  \includegraphics[width=\linewidth]{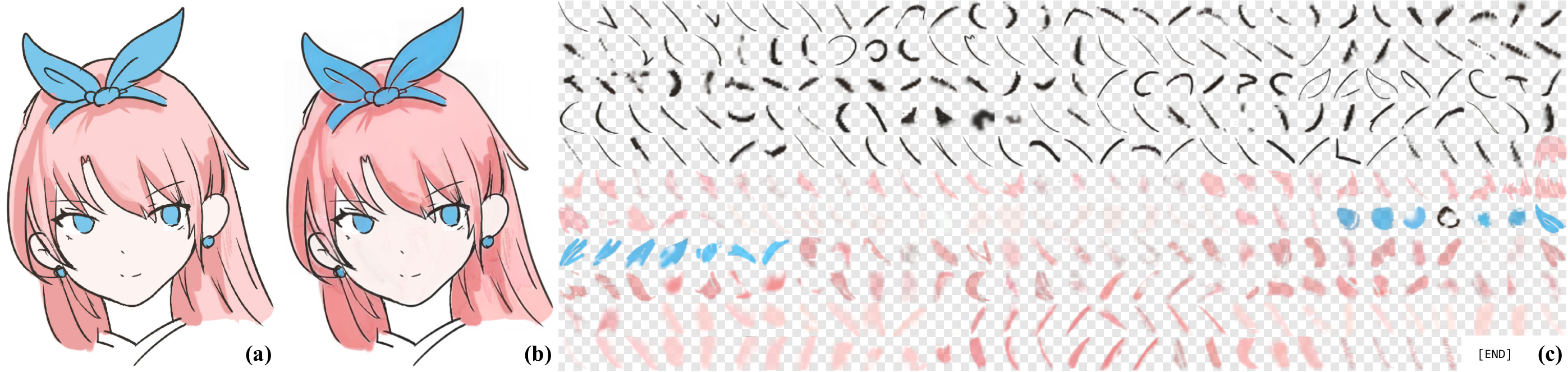}
  \caption{Reconstruction of real-life incremental drawing from timelapse videos. (a) Timelapse snapshot at $t=328$; (b) Reconstructed canvas composited with hyperstrokes; (c) Inferred stroke sequences from adjacent timelapse frames.}
  \label{fig:reconstruction}
\end{figure*}

\begin{figure}[t]
  \centering
  \includegraphics[width=\linewidth]{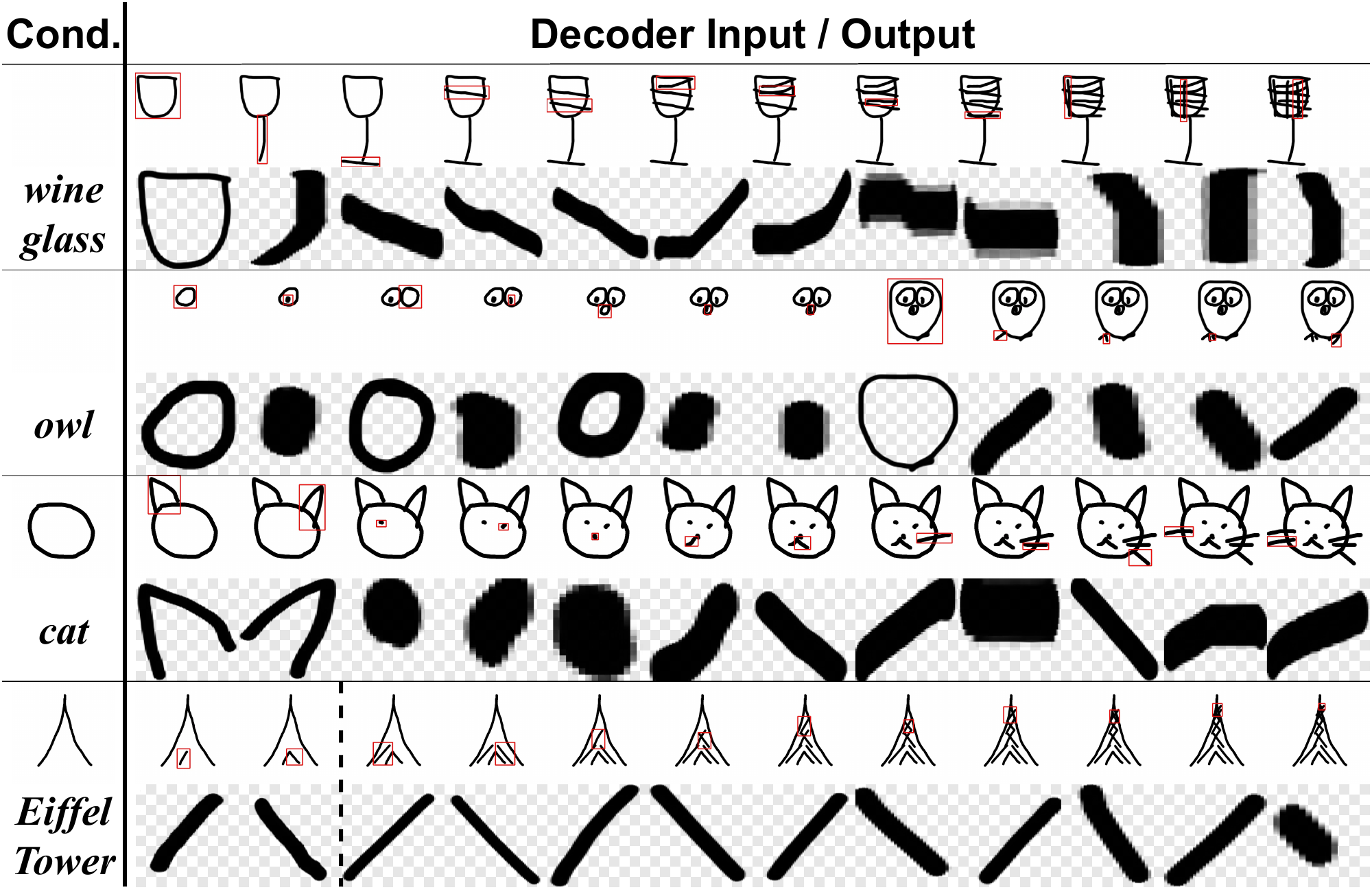}
  \caption{Results on predictive incremental drawing conditioned on raster canvas and text descriptions. Odd rows show predicted compositions; even rows demonstrate decoded grounded strokes within its bounding box. The last example prompts 2 hyperstrokes in the decoder.}
  \label{fig:gen}
\end{figure}

We choose an encoder-decoder architecture over a decoder-only model to meet the unique needs of drawing tasks. Compared with text sequences where self-attention effectively captures context, predictive drawing involves more complex contextual requirements. The focus within is to determine the next few strokes, in the context of the current canvas composition and a few past user strokes. This complexity makes a decoder-only architecture impractical, as relying on a long sequence of historical hyperstrokes would be computationally inefficient. Conversely, our encoder model $\mathcal E$ directly provides the current canvas context through a Vision Transformer, eliminating the need to learn indirectly from the complete historical hyperstroke sequences. 
This approach provides several practical applications with the context provided, including: (a) unconditional sequential hyperstroke prediction; (b) prediction of subsequent hyperstrokes using a few hyperstrokes as historical prompts, for temporal-consistent stroke prediction; and (c) predicting the next visual token $\tilde{\mathcal I}$ given a bounding box prompt. One might argue that making the decoder output a single hyperstroke would suffice, as the rasterized next-frame context could be rendered on-the-fly. However, this method fails to capture temporal information. Our approach, by predicting ordered stroke sequences, inherently captures locality of neighboring strokes, semantics of different canvas areas, as well as the drawing intent of the artists, enabling long-term understanding capability, and thus bringing better interactivity for the artists.
During training, notice that the amount of generated visual tokens \(\tilde {\mathcal I}\) and bounding box tokens \(\tilde {\mathcal B}\) are unbalanced, to stabilize the training, we further impose a coefficient \(\lambda = k / 4\) on the parts of the cross entropy loss corresponding to the generated bounding box tokens.

\section{Experiments} \label{sec:eval}

\sparagraph{Hyperstroke Representation.} We first investigate the expressiveness of the hyperstroke representation. We use our revamped VQGAN model described in Sec.\ref{subsec:hyperstroke} to reconstruct all intermediate strokes from a complete artistic drawing timelapse of 328 frames (Fig.\ref{fig:reconstruction} (a)). Figure~\ref{fig:reconstruction} (c) shows the reconstruction of strokes, grounded by their bounding box areas. The results demonstrate that tokenized hyperstroke can capture detailed stroke appearances, including shape and color variations. Based on the quality of the composition of all 328 strokes (Fig.~\ref{fig:reconstruction} (b)), we conclude that hyperstroke can successfully encode the opacity of strokes from timelapse contexts, enabling the reproduction of artistic illustrations with a much more condensed representation.

\sparagraph{Assistive Sketch Generation.} We explore the transformer model proposed in Section \ref{subsec:sequence} to predict subsequent stroke sequences from user-provided contexts. Given the challenges of scaling up the transformer to learn practical artistic drawing sequences due to the scarcity of large-scale incremental drawing datasets, we instead conduct a proof-of-concept using the \textit{Quick, Draw!} dataset \cite{ha2017neural}. Results are shown in Figure~\ref{fig:gen}. Given canvas context and text conditions, the model demonstrates the capability to generate visually pleasing, temporally intuitive, and coherent sketching sequences that compose meaningful doodles. This generation can be performed unconditionally or prompted from additional user-provided hyperstrokes.

\section{Conclusion} \label{sec:conclusion}

In this work, we propose hyperstroke, an efficient and expressive stroke representation designed to capture the essence of artistic drawing strokes. It is particularly well-suited for transformer-based sequential modeling. In the future, we may aim to investigate better hyperstroke encoding schemes, the balance between canvas encodings and historic stroke inputs, and conduct more comprehensive assistive drawing evaluations, by which we believe that the representational capabilities of hyperstroke will inspire future HCI applications in assistive drawing. It will enable a more comprehensive understanding of artistic drawing techniques and fulfill the genuine needs of artists, thereby enhancing their productivity.

\balance
\bibliographystyle{plain}
\bibliography{citation}

\begin{thebibliography}{10}

\bibitem{bhunia2020pixelor}
Ayan~Kumar Bhunia, Ayan Das, Umar~Riaz Muhammad, Yongxin Yang, Timothy~M Hospedales, Tao Xiang, Yulia Gryaditskaya, and Yi-Zhe Song.
\newblock Pixelor: A competitive sketching ai agent. so you think you can sketch?
\newblock {\em ACM Transactions on Graphics (TOG)}, 39(6):1--15, 2020.

\bibitem{dosovitskiy2020image}
Alexey Dosovitskiy, Lucas Beyer, Alexander Kolesnikov, Dirk Weissenborn, Xiaohua Zhai, Thomas Unterthiner, Mostafa Dehghani, Matthias Minderer, Georg Heigold, Sylvain Gelly, et~al.
\newblock An image is worth 16x16 words: Transformers for image recognition at scale.
\newblock {\em arXiv preprint arXiv:2010.11929}, 2020.

\bibitem{esser2021taming}
Patrick Esser, Robin Rombach, and Bjorn Ommer.
\newblock Taming transformers for high-resolution image synthesis.
\newblock In {\em Proceedings of the IEEE/CVF conference on computer vision and pattern recognition}, pages 12873--12883, 2021.

\bibitem{ha2017neural}
David Ha and Douglas Eck.
\newblock A neural representation of sketch drawings.
\newblock {\em arXiv preprint arXiv:1704.03477}, 2017.

\bibitem{huang2019learning}
Zhewei Huang, Wen Heng, and Shuchang Zhou.
\newblock Learning to paint with model-based deep reinforcement learning.
\newblock In {\em Proceedings of the IEEE/CVF international conference on computer vision}, pages 8709--8718, 2019.

\bibitem{liu2021paint}
Songhua Liu, Tianwei Lin, Dongliang He, Fu~Li, Ruifeng Deng, Xin Li, Errui Ding, and Hao Wang.
\newblock Paint transformer: Feed forward neural painting with stroke prediction.
\newblock In {\em Proceedings of the IEEE/CVF international conference on computer vision}, pages 6598--6607, 2021.

\bibitem{nitzan2024lazy}
Yotam Nitzan, Zongze Wu, Richard Zhang, Eli Shechtman, Daniel Cohen-Or, Taesung Park, and Micha{\"e}l Gharbi.
\newblock Lazy diffusion transformer for interactive image editing.
\newblock {\em arXiv preprint arXiv:2404.12382}, 2024.

\bibitem{radford2021learning}
Alec Radford, Jong~Wook Kim, Chris Hallacy, Aditya Ramesh, Gabriel Goh, Sandhini Agarwal, Girish Sastry, Amanda Askell, Pamela Mishkin, Jack Clark, et~al.
\newblock Learning transferable visual models from natural language supervision.
\newblock In {\em International conference on machine learning}, pages 8748--8763. PMLR, 2021.

\bibitem{radford2019language}
Alec Radford, Jeffrey Wu, Rewon Child, David Luan, Dario Amodei, Ilya Sutskever, et~al.
\newblock Language models are unsupervised multitask learners.
\newblock {\em OpenAI blog}, 1(8):9, 2019.

\bibitem{rombach2022high}
Robin Rombach, Andreas Blattmann, Dominik Lorenz, Patrick Esser, and Bj{\"o}rn Ommer.
\newblock High-resolution image synthesis with latent diffusion models.
\newblock In {\em Proceedings of the IEEE/CVF conference on computer vision and pattern recognition}, pages 10684--10695, 2022.

\bibitem{singh2021intelli}
Jaskirat Singh, Cameron Smith, Jose Echevarria, and Liang Zheng.
\newblock Intelli-paint: Towards developing human-like painting agents.
\newblock {\em arXiv preprint arXiv:2112.08930}, 2021.

\bibitem{zhang2018unreasonable}
Richard Zhang, Phillip Isola, Alexei~A Efros, Eli Shechtman, and Oliver Wang.
\newblock The unreasonable effectiveness of deep features as a perceptual metric.
\newblock In {\em Proceedings of the IEEE conference on computer vision and pattern recognition}, pages 586--595, 2018.

\bibitem{zheng2018strokenet}
Ningyuan Zheng, Yifan Jiang, and Dingjiang Huang.
\newblock Strokenet: A neural painting environment.
\newblock In {\em International Conference on Learning Representations}, 2018.

\end{thebibliography}

\clearpage
\nobalance
\appendix

\section{Hyperstroke Dataset Showcase}

\begin{figure}[!t]
    \centering
    \includegraphics[width=\linewidth]{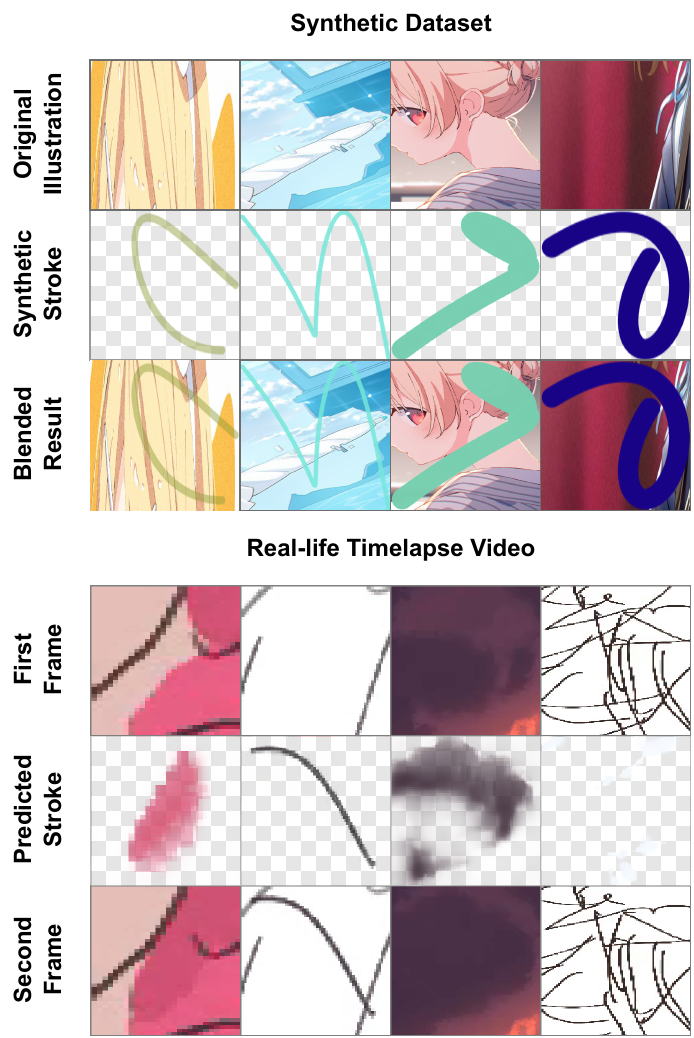}
    \caption{Data examples to train the Hyperstroke representation. The first group shows the data from synthetic dataset. From top to bottom are original illustrations, synthetic stroke images, and blended results. The supervision is conducted directly by the ground truth synthetic stroke. The second group demonstrates the data from real-life timelapse video, showing the previous frames in the frame pairs, the \textit{predicted} stroke by our model (not part of the dataset), and the latter frames in the frame pairs, from the top to bottom accordingly. Here, the supervision is implicitly applied by the two frames.}
    \label{fig:dataset}
\end{figure}

As stated, the training data for hyperstroke consists of a synthetic dataset and data from real-life timelapse video. Figure~\ref{fig:dataset} shows some samples from our dataset.

Specifically, for the synthetic dataset, we left 30\% of the strokes opaque while the other strokes have a uniform opacity sampled from $0.1$ to $1.0$.

\section{Training Details}

\subsection{Hyperstroke Representation}

Our VQ model employs a codebook of 8192 vocabularies, each with 256 dimensions of embedding. We trained the model using the mixed dataset consisting of 85,425 synthetic data samples and 74,286 real data samples in the form of frame pairs on a resolution of \(256 \times 256\). The base learning rate is \(4.5 \times 10^{-6}\), and is warmed up for 200 steps linearly at the beginning. The model is trained for 20 epochs on 4 NVIDIA A6000 GPUs for 35 hours at a total batch size of 32.

\subsection{Assistive Sketch Generation}

We evaluated the assistive sketch generation task on the \textit{Quick, Draw!} dataset~\cite{ha2017neural}, which consists of temporal vector sketching of 345 categories. We first filter the dataset for sketches with stroke number ranging from 3 to 15, subsample the dataset by \(1/5\), and then render each sketch in black color with random stroke width on arbitrary canvas positions, resulting the final dataset with 43,776,398 strokes from 7,049,475 sketches. To employ the transformer model on the new dataset, a new VQ model is trained. Under this setting, we make the following changes: we adopt the codebook size of 2048, chose \(2\times 10^{-7}\) as the base learning rate, and trained the model on a resolution of \(128 \times 128\) for 2 epochs on 8 NVIDIA A100 GPUs for 35 hours at a total batch size of 1024. The downsampling factor of the VQ model is \(16\times\), and therefore a hyperstroke consists of \(4 + (128 / 16)^2 = 68\) tokens. For the transformer model, we adopt GPT-2 (345M) \cite{radford2019language} as the decoder, a pretrained Vision Transformer (ViT)\footnote{\url{https://huggingface.co/google/vit-base-patch16-224-in21k}} as the canvas encoder, and a pretrained CLIP model\footnote{\url{https://huggingface.co/openai/clip-vit-base-patch32}} as the control encoder. Here, we condition the generation based on the category text of each sketch, and the context length is 12 strokes, i.e. \(1 + 12 \times 68 = 817\) tokens. The learning rate is \(5 \times 10^{-4}\) and we employed learning warmup and annealing. We freeze the weights of encoders during the training, and the model is trained for 1 epoch on 8 NVIDIA A100 GPUs for 3 days at a total batch size of 1024.

\section{Additional Experiment Results}

\subsection{Hyperstroke Reconstruction}

\begin{figure*}
    \centering
    \includegraphics[width=\linewidth]{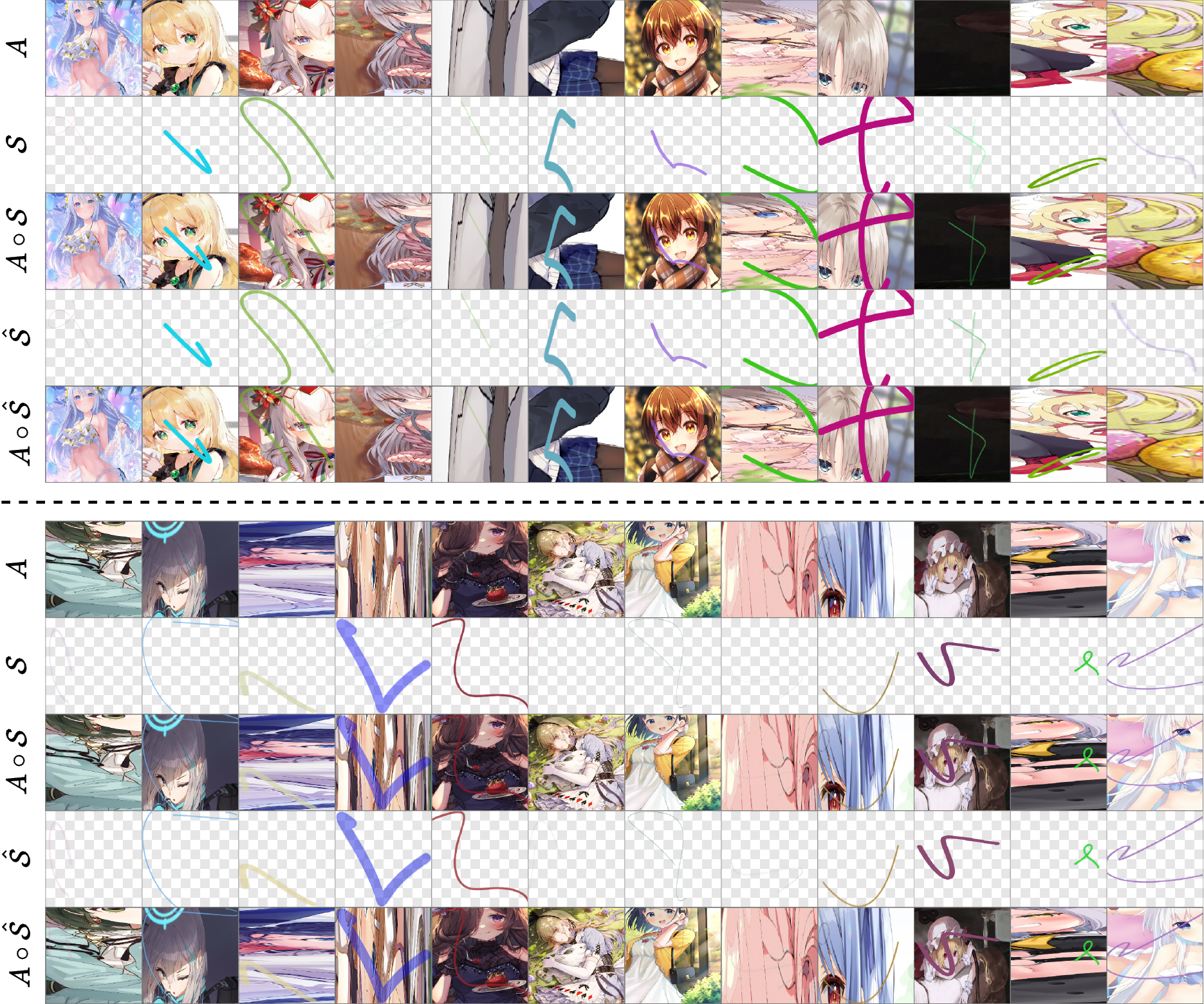}
    \caption{Hyperstroke model result on synthetic dataset. For each group, the five rows from the top to the bottom stand for the original cropped illustration, the generated ground truth stroke images, the blended illustration by the ground truth, the predicted strokes between the two frames, and finally the blended result of the predicted strokes.}
    \label{fig:hyperstroke-more-synthetic}
\end{figure*}

\begin{figure*}
    \centering
    \includegraphics[width=\linewidth]{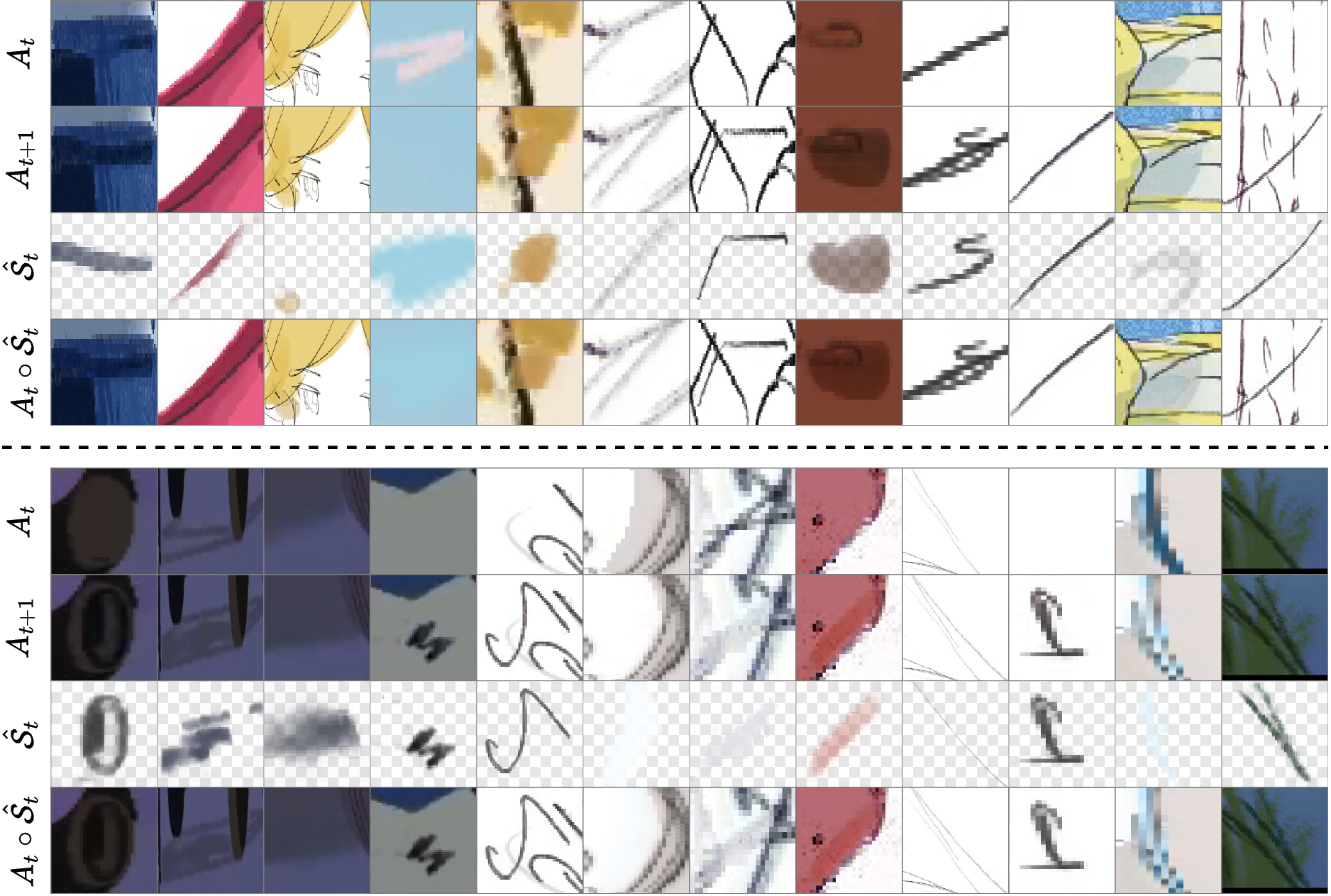}
    \caption{Hyperstroke model result on real-life timelapse drawing data. For each group, the four rows from the top to the bottom stand for the previous frame, the latter frame, the predicted strokes between the two frames, and finally the blended result of the predicted strokes onto the initial frames.}
    \label{fig:hyperstroke-more-timelapse}
\end{figure*}

Figure~\ref{fig:hyperstroke-more-synthetic} and Figure~\ref{fig:hyperstroke-more-timelapse} show results on synthetic dataset and real-life drawing timelapse data accordingly.

\subsection{Assistive Sketch Generation}

\begin{figure*}
    \centering
    \includegraphics[width=\linewidth]{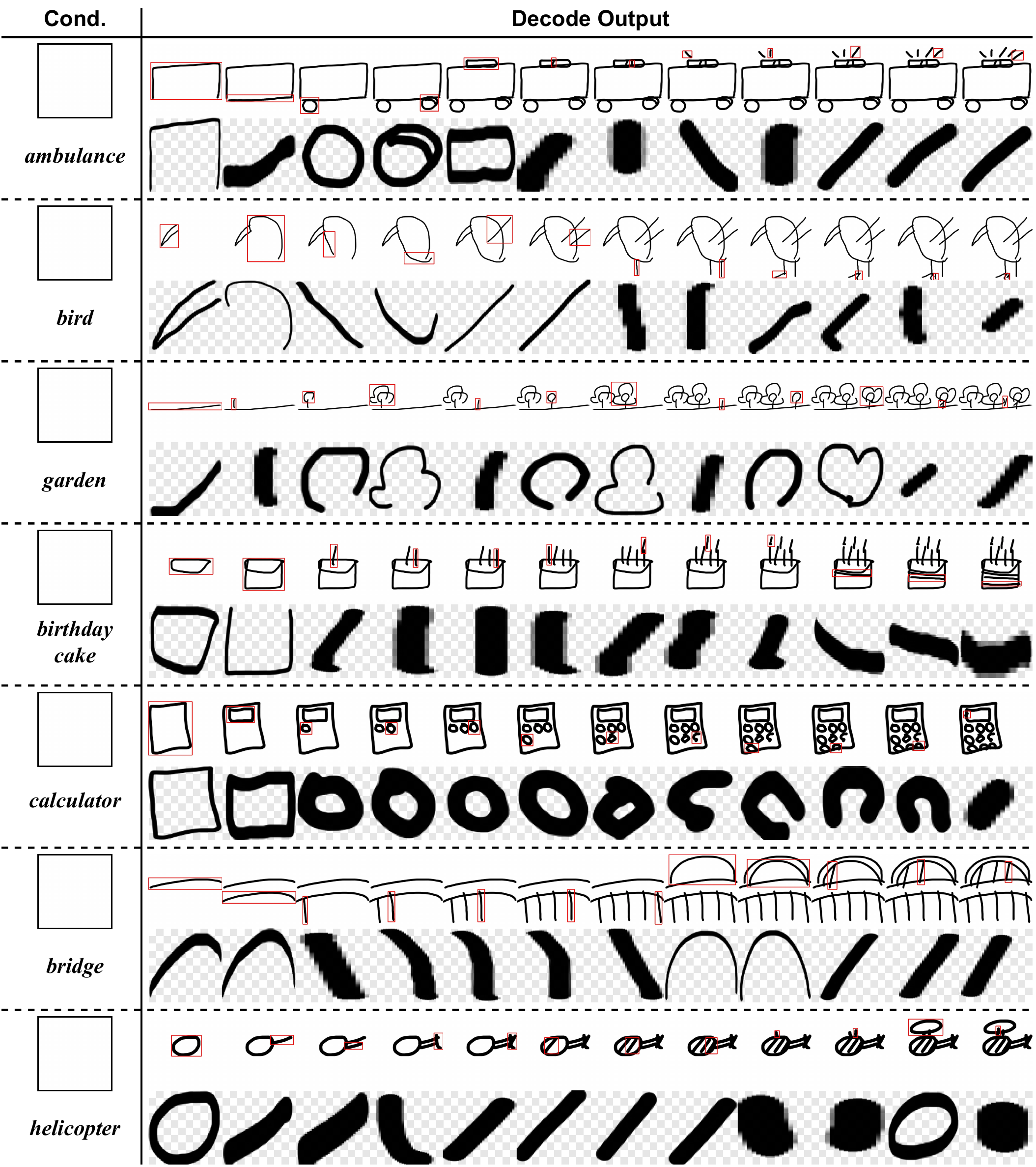}
    \caption{Results on assistive sketch generation from blank canvas, conditioned on seen text categories.}
    \label{fig:gen-more-blank}
\end{figure*}

\begin{figure*}
    \centering
    \includegraphics[width=\linewidth]{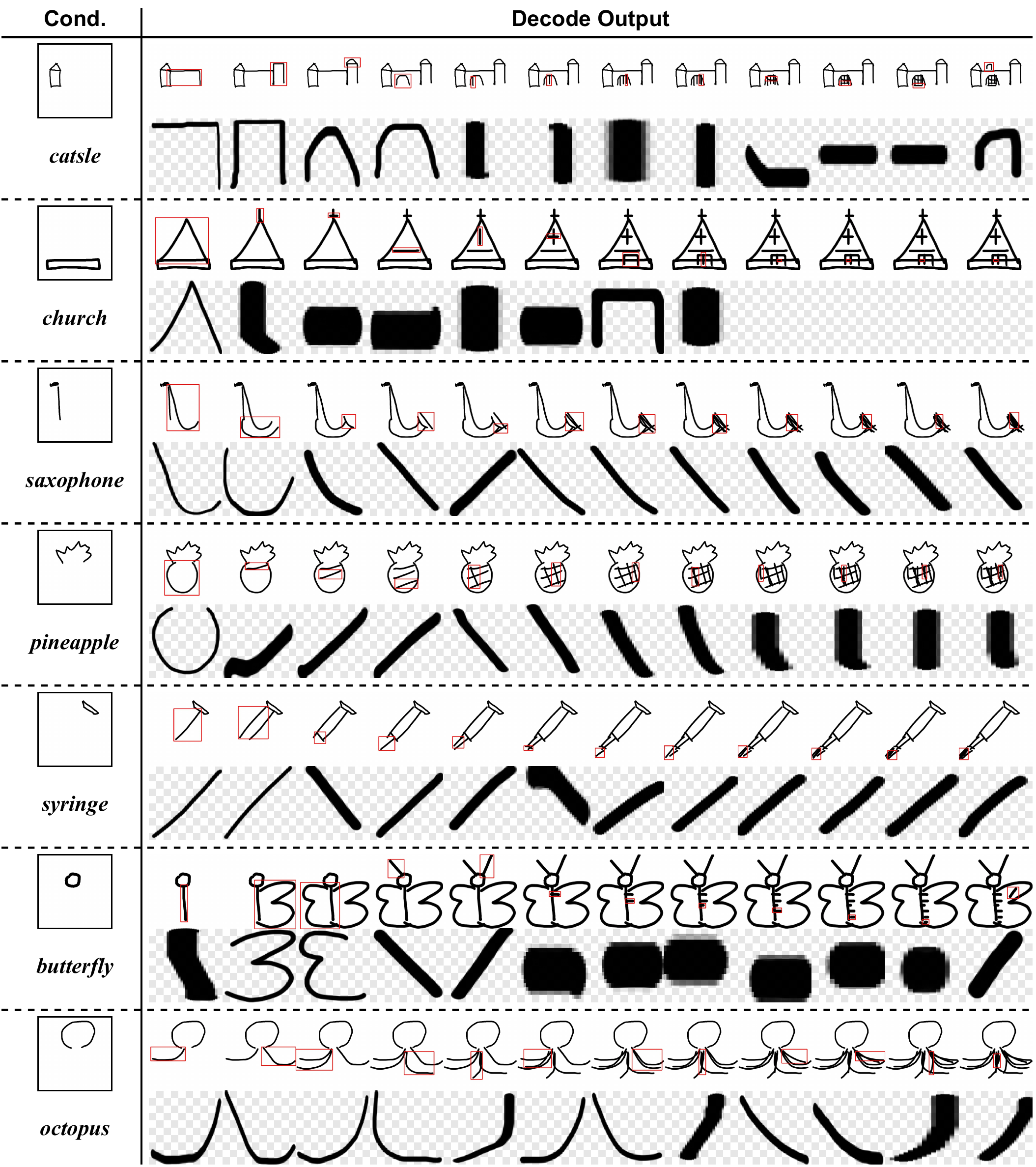}
    \caption{Results on assistive sketch generation, conditioned the raster canvas images and seen text categories.}
    \label{fig:gen-more-completion}
\end{figure*}

\begin{figure*}
    \centering
    \includegraphics[width=\linewidth]{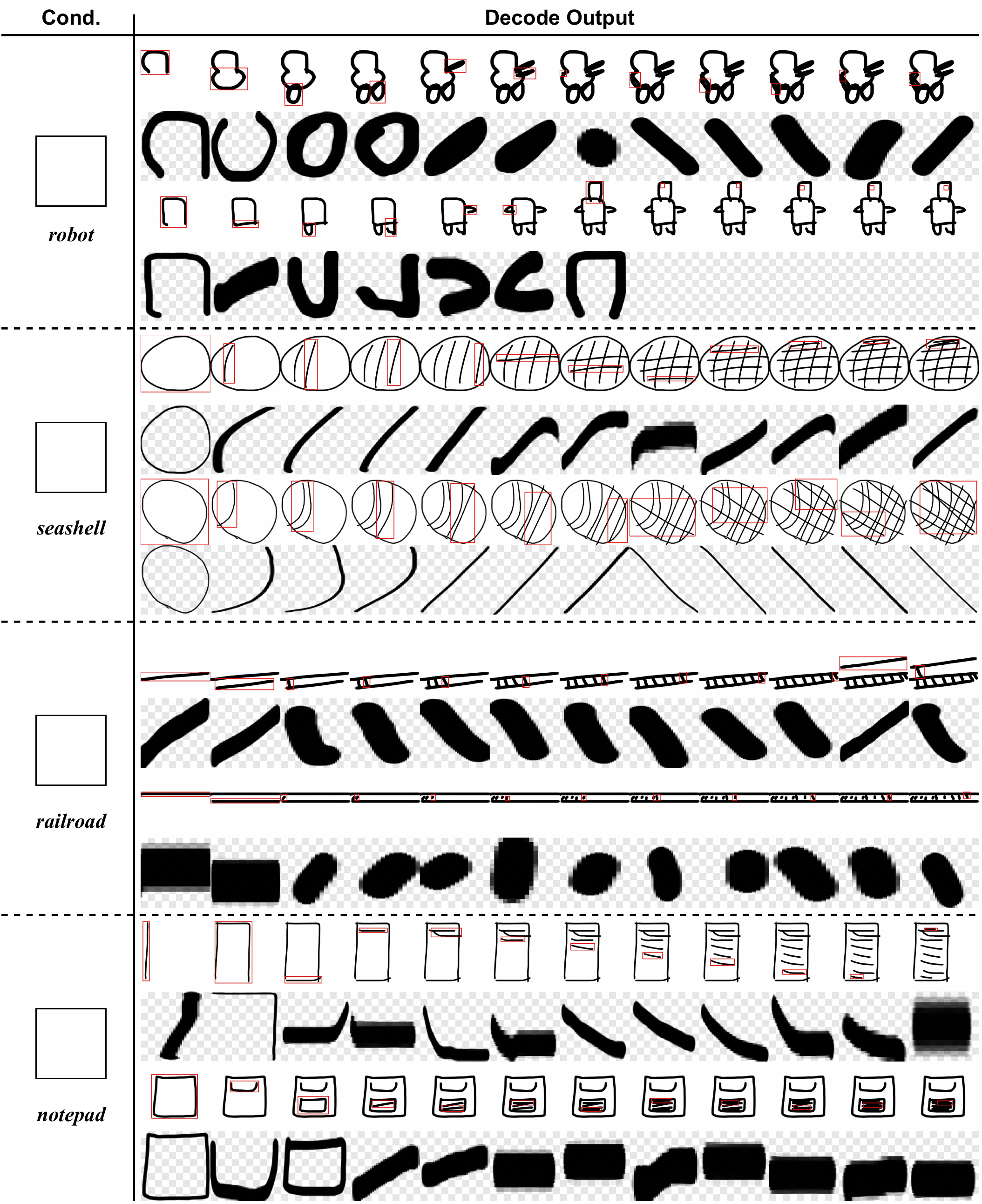}
    \vspace{-1em}
    \caption{Results on assistive sketch generation from blank canvas, conditioned on unseen text categories.}
    \label{fig:gen-more-blank-unseen}
\end{figure*}

Figure~\ref{fig:gen-more-blank} shows generation results conditioned on blank canvas and seen text categories. Figure~\ref{fig:gen-more-completion} demonstrates the results where the canvas is half-way finished. We also tested our model on \textit{unseen} text conditions beyond the 345 text categories the model is trained on as shown in Figure~\ref{fig:gen-more-blank-unseen}. The model demonstrates the capability of extrapolation to some extent, where it can guess the overall shape and feel of unseen data in some cases.

\balance

\end{document}